\documentclass[final]{vgtc}                          

\usepackage{microtype}  
\usepackage{graphicx}
\usepackage{textcomp}               
\usepackage{mathptmx}                 
\usepackage{times}                    
\usepackage{cite}                     
\usepackage{tabu}  
\usepackage{booktabs}                
\usepackage{enumitem}
\usepackage{mathptmx}
\usepackage{microtype}
\usepackage{xargs}                      
 
 \usepackage[table,xcdraw]{xcolor}

\title{Integrating Eye-Gaze Data into CXR DL Approaches: A Preliminary study}

\author{Andr\'{e} Lu\'{i}s 
     \scriptsize INESC-ID / Instituto Superior T\'{e}cnico \\
     \scriptsize Universidade de Lisboa, Portugal\\
     \scriptsize andre.t.luis@tecnico.ulisboa.pt\\
       \and Chihcheng Hsieh $^*$\\ 
      \scriptsize School of Information Systems \\
      \scriptsize Queensland University of Technology, Australia \\
      \scriptsize chihcheng.hsieh@hdr.qut.edu.au \\
    \and Isabel Blanco Nobre\\
       \scriptsize  Imagiology Department\\ 
        \scriptsize  Grupo Lusíadas, Lisboa, Portugal\\
        \scriptsize isabel.blanco.nobre@lusiadas.pt\\
    \and Sandra Costa Sousa \\
 \scriptsize Imagiology Department\\
  \scriptsize  Grupo Lusíadas, Lisboa, Portugal \\
  \scriptsize sandra.costa.sousa@lusiadas.pt
     \and Anderson Maciel\\ 
        \scriptsize INESC-ID / Instituto Superior T\'{e}cnico \\
        \scriptsize Universidade de Lisboa, Portugal \\
        \scriptsize anderson.maciel@tecnico.ulisboa.pt \\
         \and Joaquim Jorge\\
     \scriptsize INESC-ID / Instituto Superior T\'{e}cnico \\
     \scriptsize Universidade de Lisboa, Portugal\\
     \scriptsize jorgej@acm.org \\
     \and Catarina Moreira\\ 
        \scriptsize INESC-ID / School of Information Systems \\
        \scriptsize Queensland University of Technology \\
        \scriptsize catarina.pintomoreira@qut.edu.au
    }

\abstract{

This paper proposes a novel multimodal DL architecture incorporating medical images and eye-tracking data for abnormality detection in chest x-rays. Our results show that applying eye gaze data directly into DL architectures does not show superior predictive performance in abnormality detection chest X-rays. These results support other works in the literature and suggest that human-generated data, such as eye gaze, needs a more thorough investigation before being applied to DL architectures.

}

\begin{document}

\firstsection{Introduction}

\maketitle

\begin{figure*}[!t]
\setlength\abovecaptionskip{-0.1\baselineskip}
    \resizebox{2.10\columnwidth}{!}{
    \includegraphics{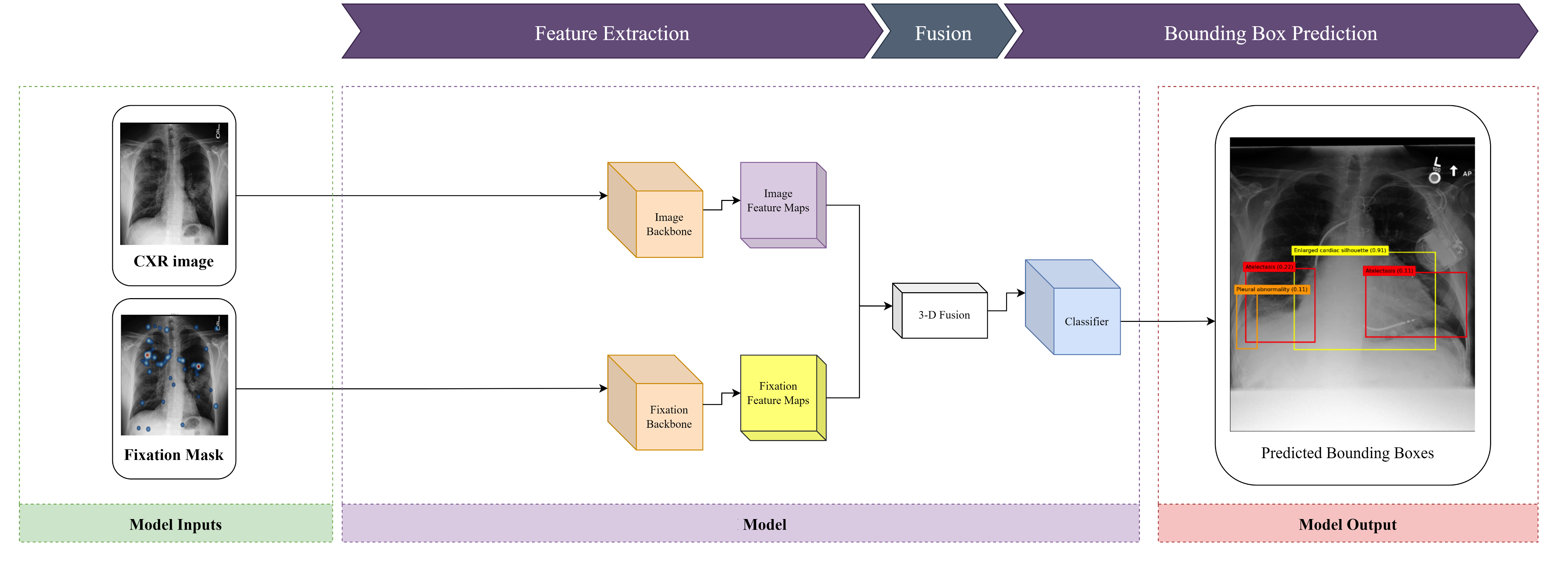}
    }
    \caption{Proposed Multimodal DL architecture that combines fixation maps with chest x-ray images and performs abnormality detection for five classes: enlarged cardiac silhouette, atelectasis, pleural abnormality, consolidation, and pulmonary edema
    }
    \label{fig:multimoal_arch}
\end{figure*}


Diagnostic imaging plays a crucial role in diagnosing and treating a wide range of diseases, including cancer, heart disease, and injuries. It enables medical professionals to make accurate and timely diagnoses and determine the most effective treatment plans. However, access to radiologists, especially in underserved and remote areas, remains a significant challenge. The shortage of radiologists worldwide and the difficulty of providing radiology services to remote locations contribute to this issue. This can lead to delays in diagnosis, increased medical costs, and poorer patient outcomes.


Deep learning (DL) has emerged as a powerful solution to the challenges of diagnostic imaging in underserved and remote areas. DL algorithms are able to extract and learn complex features from medical images with high accuracy, enabling them to automatically analyze images and identify patterns and abnormalities that may be difficult for human radiologists to detect. This can help to reduce the dependence on radiologists and improve access to healthcare for patients in underserved areas. However, these systems are also subject to bias and lack transparency, which can lead to a decline in radiologists' trust and adoption in healthcare. 


Virtual reality (VR) technology has the potential to transform the way radiologists assess medical images, providing a game-changing solution to the challenges of radiologist shortages and access to healthcare \cite{sousa17vrrroom}. By creating an immersive, virtual radiology reading room, VR enables radiologists to work remotely and provide accurate diagnoses even in the face of radiologist shortages. This is particularly significant during pandemics, where healthy radiologists may have to stay home, and diagnoses may be delayed due to radiologist shortages. Furthermore, VR technology can also open new research paths for DL approaches by incorporating new data modalities, such as eye tracking. Eye tracking in VR has been shown to be able to estimate cognitive load \cite{Rayner2015EvidenceCongnitiveFixation,tolvanen2022}, fatigue \cite{Reichle2012EyeMindLink1} and detect true/false positives \cite{Mall2017}. By providing valuable information about how radiologists interact with medical images \cite{leveque2018}, VR-based eye tracking can help improve the diagnostic accuracy of DL systems \cite{Moreira2022xr}.

In this article, we present the initial findings of a cutting-edge multimodal deep learning (DL) architecture that integrates medical images and eye-tracking as a novel modality for detecting abnormalities in chest x-rays. The goal of this research is to evaluate the potential of this approach to enhance the diagnostic accuracy of DL systems. Eye-tracking, the process of measuring the eye gaze, can provide valuable insights into the diagnostic process, such as where the radiologist is focusing and how they are interpreting the image. By incorporating this data into our DL architecture, we aim to reduce the dependence on radiologists and improve access to healthcare for patients in underserved areas. Additionally, we aim to investigate whether incorporating eye-tracking can make DL systems more robust to biases and improve their diagnostic accuracy.

\section{Related Work}

Diagnostic imaging is a well-established field with a long history of research aimed at improving the accuracy and accessibility of medical imaging. Although there has been a growing body of research on the use of DL for medical imaging diagnostics \cite{CALLI2021102125}, these studies still suffer from the detection of spurious correlations that result in biased and erroneous diagnosis. This includes studies on the use of DL for detecting abnormalities in medical images such as X-ray, CT and MRI scans \cite{Moreira2020medical}. The potential solution to this problem involves incorporating multiple data modalities into DL architectures to reduce reliance on medical images alone in the learning process.



Previous research has investigated the integration of eye-tracking data with medical images, specifically for training purposes \cite{Kosel2021}. More recent works have shifted the focus towards exploring the implications of eye-tracking in deep learning \cite{Mall2017}. The EYEGAZE dataset~\cite{karargyris2021paper} and REFLACX~\cite{Bigolin2022} are two examples of such datasets. The Eye-Gaze dataset contains 1,083 frontal chest x-ray (CXR) image readings by one radiologist, as well as dictations reports. The images used were sourced from the MIMIC database~\cite{Johnson2019}, and image-level labels were extracted from MIMIC reports using natural language processing techniques. REFLACX, on the other hand, contains 3,032 synchronized sets of eye-tracking data and timestamped report transcriptions for images from the MIMIC-CXR dataset. In this case, CXRs were annotated by five radiologists, who identified bounding boxes and their labels on each image. These datasets provide valuable resources for investigating the use of eye-tracking data in combination with medical images to improve diagnostic accuracy. The public availability of these two eye-tracking datasets for medical images facilitated the investigation of new multimodal DL architectures that integrate information from radiologists' eye tracking data with the patients' medical images.


Karargyris et al. \cite{karargyris2021paper} proposed two DL architectures to incorporate eye-tracking data in predicting image-level labels for chest x-ray (CXR) images. The first approach concatenates CXR images passed through a convolutional neural network (CNN) with temporal fixation heatmaps through a 1-layer bidirectional long short-term memory network with self-attention \cite{Sepp1997}. The results showed a 5\% area under the curve (AUC) improvement using temporal fixation heatmaps, compared with the baseline model using only CXR image data as input. The second approach used static fixation heatmaps, i.e., aggregating all the temporal fixations in a single image. During training, the model jointly trains the static fixation heatmap and the image-level label. During testing, when a CXR image is given as input, the model outputs the label and a heatmap distribution of the most important locations of the condition. This approach showed similar results to the baseline model but with added interpretability. 

Gaze data in DL were also studied by Wang et al. \cite{wang2022follow}, who collected gaze information from radiologists and modelled a gaze-guided attention network, which ensures the network can focus on the disease regions like the radiologists and outputs the disease label, as well as an attention map based on the annotated bounding boxes and the gaze information from radiologists. The results show that the use of the radiologist's gaze as supervision in the GA-net architecture outperformed state-of-the-art methods using only images, such as ResNet \cite{Woo2018}, or the Vision Transformer \cite{Dosovitskiy2020}. Also, the collection of gaze data compared to manually annotated bounding boxes from radiologists showed to be faster and led to a similar performance in classification accuracy. Other works suggest that eye tracking can be useful since the first fixations that the radiologist makes usually coincide with regions containing some lesions \cite{Nodine1987}.


Despite the potential benefits of incorporating eye-tracking data into DL models for diagnostic imaging, several challenges have been identified in the literature. Some studies have suggested that saliency maps from human fixations should not be used in DL, as there is evidence that these systems may use background context for object classification, leading to biases \cite{Qi2019,Nie2018}. Additionally, one of the first studies to use eye-tracking data in CXR analysis \cite{Carmody1981} distinguished three types of diagnostic errors using eye-tracking data: (1) search errors, which occur when the target is missed; (2) recognition errors, which happen when the eyes fixate on the target, but the target is still missed; and (3) decision errors, which relate to the inability of the radiologist to report the findings. These errors can contaminate the collection of eye-tracking data and negatively affect the performance of DL models. Studies have reported that the proportion of these errors can be as high as 30\% search errors, 25\% recognition errors, and 45\% decision errors \cite{Krupinski1996VisualSP,HU199425}. These challenges must be taken into consideration when designing and evaluating DL models that incorporate eye-tracking data to improve diagnostic accuracy. It is the purpose of this study to further examine whether DL technologies can benefit from eye-tracking data in medical diagnosis.



\section{A Multimodal Deep Learning Architecture with Fixation Masks}

The proposed architecture is an extension of the Mask-RCNN network \cite{Girshick2015}. Mask R-CNN is a state-of-the-art algorithm for object detection and instance segmentation in images and video frames. It uses a region proposal network (RPN) to generate potential regions in an image that may contain objects and then uses a separate network to classify and segment the objects within those regions. 

The key innovation of Mask R-CNN is the addition of a fully convolutional network branch on top of the RPN and classification network, which generates a binary mask for each instance of an object in the image. This allows for more precise localization of objects within an image, as it can identify the specific pixels that belong to each object.

The architecture of Mask R-CNN consists of three main components: \textit{the backbone network}, which is used to extract feature maps from the input image (typically a pre-trained convolutional neural network (CNN) such as ResNet); \textit{the RPN} that generates object proposals from the feature maps; and \textit{the detection network} that classifies and segments the objects within the proposals.


We extended this architecture by adding a new backbone based on the fixation maps of the radiologists' eye gaze patterns. We used  Mobilenet as a backbone since it performs well on small datasets \cite{Howard2017}. Figure~\ref{fig:multimoal_arch} presents the proposed multimodal model architecture. First, the CXR image and fixation mask are fused using an element-wise sum. Then, the RPN outputs a set of rectangular object proposals, and the pooling layer aligns the regions of interest with the input into a single feature map. Finally, these are flattened and passed to the classifier, which outputs, for each candidate region, the bounding box coordinates, a class label, and the object mask.

The inner workings and outputs of the multimodal approach are similar to the baseline technique. A key difference lies in the input. Instead of a single CXR image, both the image and the fixation mask, containing the heatmap of a radiologist's fixations during a reading, serve as input. The fusion between the two is achieved through element-wise multiplication, where the features of both image representations are multiplied together, element by element.
\begin{equation}
\label{eqn:6.1}
L = L_{classification} + L_{bbox} + L_{mask}
\end{equation}
The loss function in the Mask R-CNN architecture for each sample's region of interest is represented in equation \ref{eqn:6.1}. The classification and bounding box losses are the same as in Faster R-CNN \cite{Girshick2015}, which means that the classification loss corresponds to the binary cross-entropy (object vs. not object). In contrast, the regression loss (bounding box) is the smooth L1 norm. The mask loss is only defined for the ground truth class and is the average binary cross-entropy loss. During training, these losses were monitored throughout the epochs, where a decreasing pattern indicates that meaningful relations are being learned.

During the evaluation, the Intersection over the detected B-Box area ratio (IoBB) threshold is used to determine the correctness of predictions. A prediction is considered correct (true positive) only if it intersects the ground truth box above the IoBB threshold. In order to observe the models' ability to localise, we also test their performance on different IoBB thresholds. In the medical field, the IoBB is more preferable than the Intersection over Union ratio (IoU) since the ground truth boxes are often oversized in order to include all the scattered or large lesions. To evaluate the different models, absolute precision and recall were used, for a 50\% intersection over bounding boxes. While precision refers to the proportion of lesion class predictions that belong to the lesion class, recall concerns the proportion of lesion class predictions made from all lesion examples in the dataset. Training these architectures does not start from scratch but rather from pre-trained models on publicly available datasets. Here, we tested two backbones, 
ResNet \cite{He2015} and Mobilenet \cite{Howard2017}, both pre-trained on Microsoft COCO dataset \cite{Lin2014COCO}. 

\section{Dataset}

REFLACX \cite{Bigolin2022} is a dataset of reports and eye-tracking data for localization of abnormalities in 3,032 CXR images. It is one of the first publicly available datasets containing eye gaze location, pupil data, and dictations from radiologists while performing the CXR image readings. Besides, the abnormality bounding boxes were identified by five radiologists, acting as local labels that can be further used for object localization. 

The images shown to the radiologists in the data collection procedures were randomly sampled from MIMIC-CXR dataset \cite{Johnson2019}. After sampling, outlier images were excluded, either due to missing parts or because they were flipped. 

\begin{figure}[!b]
    \setlength\abovecaptionskip{-0.05\baselineskip}
    \resizebox{\columnwidth}{!}{
    \includegraphics[scale=0.2]{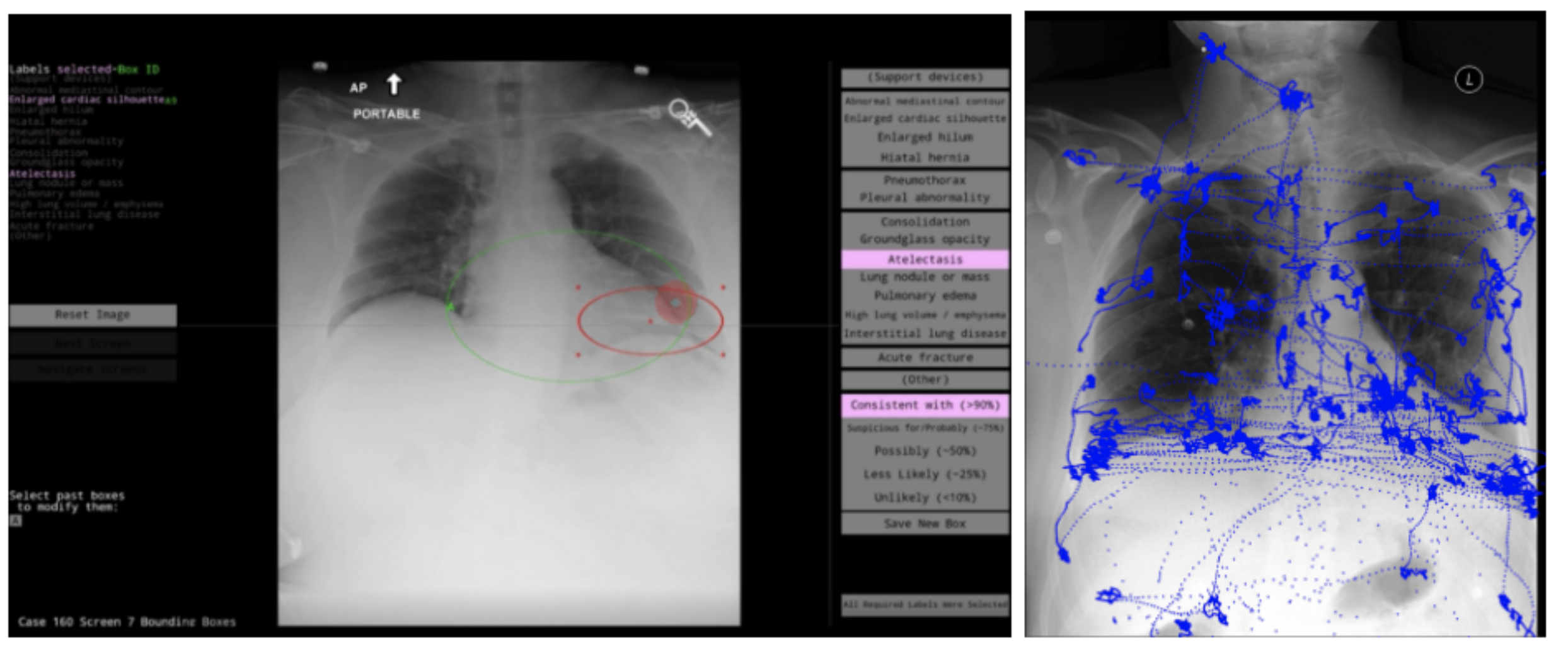}
    }
    \caption{Example of the REFLACX data collection interface (left) and the recorded eye gaze data from the radiologist.}
    \label{fig:REFLACX}
\end{figure}

 A CXR reading refers to a single data collection session where one radiologist analyses one CXR image from one patient. In the session, the professional can identify zero, one or more abnormality location ellipses. Figure \ref{fig:REFLACX} shows an example of the interface used to collect the data \cite{Bigolin2022} and the radiologist's eye gaze patterns. 
 
\section{Results and Discussion}
We compared the Mark-RCNN using only a CXR (the baseline) with the proposed multimodal DL architecture that combines both CXRs and fixation maps. Table~\ref{tab:results} presents the obtained results. 


\begin{table*}[!ht]
\resizebox{2\columnwidth}{!}{
\begin{tabular}{lcccc}
\cline{2-5}
\multicolumn{1}{l|}{} &
  \multicolumn{2}{c|}{\cellcolor[HTML]{2980B9}{\color[HTML]{FFFFFF} \textbf{Mask CNN (image only)}}} &
  \multicolumn{2}{c|}{\cellcolor[HTML]{2980B9}{\color[HTML]{FFFFFF} \textbf{Mask\_CNN (images + fixations)}}} \\ \hline
\rowcolor[HTML]{FFFFFF} 
\multicolumn{1}{|l|}{\cellcolor[HTML]{2980B9}{\color[HTML]{FFFFFF} \textbf{Abnormality}}} &
  \textbf{AP@{[}IoBB=0.50{]}} &
  \multicolumn{1}{c|}{\cellcolor[HTML]{FFFFFF}\textbf{AR@{[}IoBB=0.50{]}}} &
  \textbf{AP@{[}IoBB=0.50{]}} &
  \multicolumn{1}{c|}{\cellcolor[HTML]{FFFFFF}\textbf{AR@{[}IoBB=0.50{]}}} \\ \hline
\rowcolor[HTML]{FFFFFF} 
\multicolumn{1}{|l|}{\cellcolor[HTML]{FFFFFF}\textbf{Enlarged Cardiac Silhouette}} &
  0.429326 &
  \multicolumn{1}{c|}{\cellcolor[HTML]{FFFFFF}0.810000} &
  0.436229 &
  \multicolumn{1}{c|}{\cellcolor[HTML]{FFFFFF}0.760000} \\
\rowcolor[HTML]{FFFFFF} 
\multicolumn{1}{|l|}{\cellcolor[HTML]{FFFFFF}\textbf{Atelectasis}} &
  0.125410 &
  \multicolumn{1}{c|}{\cellcolor[HTML]{FFFFFF}0.529412} &
  0.092772 &
  \multicolumn{1}{c|}{\cellcolor[HTML]{FFFFFF}0.192513} \\
\rowcolor[HTML]{FFFFFF} 
\multicolumn{1}{|l|}{\cellcolor[HTML]{FFFFFF}\textbf{Pleural abnormality}} &
  0.043422 &
  \multicolumn{1}{c|}{\cellcolor[HTML]{FFFFFF}0.226519} &
  0.052553 &
  \multicolumn{1}{c|}{\cellcolor[HTML]{FFFFFF}0.165746} \\
\rowcolor[HTML]{FFFFFF} 
\multicolumn{1}{|l|}{\cellcolor[HTML]{FFFFFF}\textbf{Consolidation}} &
   0.113867 &
  \multicolumn{1}{c|}{\cellcolor[HTML]{FFFFFF}0.308642} &
  0.010061 &
  \multicolumn{1}{c|}{\cellcolor[HTML]{FFFFFF}0.086420} \\
\rowcolor[HTML]{FFFFFF} 
\multicolumn{1}{|l|}{\cellcolor[HTML]{FFFFFF} \textbf{Pulmonary edema}} &
  0.030728 &
  \multicolumn{1}{c|}{\cellcolor[HTML]{FFFFFF}0.410959} &
  0.001856 &
  \multicolumn{1}{c|}{\cellcolor[HTML]{FFFFFF}0.027397} \\ \hline
\rowcolor[HTML]{FFFFFF} 
\multicolumn{1}{|l|}{\cellcolor[HTML]{FFFFFF}\textbf{Average}} &
  0.148551 &
  \multicolumn{1}{c|}{\cellcolor[HTML]{FFFFFF}0.457106} &
  0.246415 &
  \multicolumn{1}{c|}{\cellcolor[HTML]{FFFFFF}0.144261} \\ \hline
 &
  \multicolumn{1}{l}{} &
  \multicolumn{1}{l}{} &
  \multicolumn{1}{l}{} &
  \multicolumn{1}{l}{}
\end{tabular}
}
\vspace{-6pt}
\caption{Results obtained using the baseline model Mask-RCNN using only the chest Xray and the proposed multimodal DL architecture that combines chest Xray images with radiologists fixation maps extracted from eye gaze patterns. The results support the works of the literature that claim that applying raw eye tracking data to DL approaches brings no predictive advantage.}
\label{tab:results}
\end{table*}

The results indicate that the model using only images performs better than the multimodal approach using the radiologists' fixation masks. We believe that these poor results are due to the fact that fixation masks consist of the clustering of gaze data, which means that a CXR image will contain a small number of fixations. Those fixations mostly do not correlate with the groundtruth annotations, implying that more investigation is needed to extract regions of interest from eye gaze data that correlate with the lesion annotations in CXRs. Additionally, further analysis indicated that the REFLACX data contains a lot of noise: the eye gaze patterns that were recorded and made available to the public contained gazes resulting from the interaction with the interface and with the CXR assessment.  

When looking at the performance of the models for each abnormal condition alone, Mask-CNN shows a clear performance advantage. For this model, it is easy to identify \textit{enlarged cardiac silhouette} conditions (with an AP@[BBOX=0.5] = 43\%). The reason for this is due to its more localized nature around the heart. The remaining lesions show similar low performances, probably because they mostly occur together in the lungs region, making it difficult for the classifier to differentiate them.

Although in terms of performance, we could not prove any superior advantage of using eye trakcing data in a multimodal DL architecture, this preliminary study enabled us to extract several insights for future research:

\noindent
1. The current approaches in deep learning working on the REFLACX dataset tend to utilize raw eye gaze data without taking into account the intricacies of its human generation process. This results in a data stream that is highly noisy and does not effectively contribute to the supervised learning process.

\noindent
2. The data collection process employed in REFLACX utilized traditional desktop eye-trackers, which are known to be highly susceptible to noise and do not permit extreme head movement during data collection. This can result in missing data points or inaccuracies in the recorded data.

\noindent
3. Eye tracking data is a form of human-centric data. As such, it is essential to consider the human factors involved in its generation when processing and analyzing this data. This includes, but is not limited to, the examination of human search patterns.

\section{Conclusion and Future Directions}
Eye-tracking data have been widely studied in medical imaging. More recently, researchers have started investigating the potential of combining eye-tracking data in DL approaches. The reason for this is to develop novel DL architectures that can promote more accurate and precise medical diagnoses based on the radiologists' behavioral patterns. These systems could be a plausible solution for the current shortage of radiologists worldwide. The few works that used DL to combine eye-tracking data with CXR images show mixed results: some report predictive advantages, while others do not. 

In this work, we investigated this divergence 
by proposing a multimodal DL architecture based on Mark-RCNN that combines radiologists' fixation maps with CXR images. Our results showed that incorporating fixation maps showed no performance advantage.

For future research, we plan to extend this study to the analysis of the radiologists' pupil dilations while reading a CXR. Literature shows a strong correlation between pupil dilations and cognitive load \cite{bednarik2018}, which can be a better metric to correlate with abnormal regions in the CXR than fixation maps. 

From this study, we also verified that the desktop eye trackers used to collect data are highly susceptible to noise and head movements. For future work, we also intend to replicate the REFLACX data collection process using VR glasses. In a VR setting, conditions of light and contrast can be easily controlled, the radiologist will not be bounded to restricted head movements, and the data eye gaze data collection is less noisy.

\acknowledgments{
This 
work partially supported by the UNESCO Chair on AI\&XR; and the Portuguese \textit{Funda\c c\~ao para a Ci\^encia e a Tecnologia (FCT)} under grants no. 2022.09212.PTDC and no. UIDB/50021/2020.
}

\bibliographystyle{abbrv}
\nocite{jorge2011sketch,correia05,Pereira04,goncalves03,moreira15,moreira11}

\end{document}